\documentclass{ios}
\usepackage{epsfig}
\usepackage{url}
\usepackage{cite}

\usepackage{amsmath,amsthm,amssymb,amsfonts}
\makeatletter
\thm@headfont{\sc}
\makeatother
\newtheorem{theorem}{Theorem}

\begin{document}

\title{Kernel Regression with Sparse Metric Learning}
\author{Rongqing Huang and Shiliang Sun\thanks{Corresponding author. Tel.: +86 2154345186; fax: +86 2154345119. E-mail address: slsun@cs.ecnu.edu.cn (S. Sun).}\\
{\small\it Department of Computer Science and Technology, East China Normal University}\\
{\small\it 500 Dongchuan Road, Shanghai 200241, P. R. China}}

\ioshead{R. Huang and S. Sun}{Kernel Regression with Sparse Metric Learning}

\maketitle
\begin{abstract}
Kernel regression is a popular non-parametric fitting technique. It aims at learning a function which estimates the targets for test inputs as precise as possible. Generally, the function value for a test input is estimated by a weighted average of the surrounding training examples. The weights are typically computed by a distance-based kernel function and they strongly depend on the distances between examples. In this paper, we first review the latest developments of sparse metric learning and kernel regression. Then a novel kernel regression method involving sparse metric learning, which is called kernel regression with sparse metric learning (KR$\_$SML), is proposed. The sparse kernel regression model is established by enforcing a mixed $(2,1)$-norm regularization over the metric matrix. It learns a Mahalanobis distance metric by a gradient descent procedure, which can simultaneously conduct dimensionality reduction and lead to good prediction results. Our work is the first to combine kernel regression with sparse metric learning. To verify the effectiveness of the proposed method, it is evaluated on 19 data sets for regression. Furthermore, the new method is also applied to solving practical problems of forecasting short-term traffic flows. In the end, we compare the proposed method with other three related kernel regression methods on all test data sets under two criterions. Experimental results show that the proposed method is much more competitive. 
\end{abstract}

\vspace{11pt}                
\section{Introduction}
One of the oldest and most commonly used algorithms for regression is kernel regression. Kernel regression is a non-parametric technique rooting in statistics for estimating the conditional expectation of a random variable. In non-parametric regression, the conditional expectation of a variable $\mathbf{Y}$ given a variable $\mathbf{X}$ is written as $E(\mathbf{Y}|\mathbf{X})=m(\mathbf{X})$, where the unknown function $m$ is approximated by a locally weighted average, using a kernel as the weighting function\cite{1,2,3}. The standard regression task is to estimate an unknown function based merely on a training set of evaluations that are possibly noisy. The target value of a test input is a real number and it is computed using a weighted average of the function values observed at its \emph{k} nearest neighbors in the training set. The weight of each training point is obtained by a kernel function, which typically decays rapidly with the
distance between itself and the test point. This is also the reason why the target value of a test point has a strong dependence on its nearby training points\cite{4}.

The traditional kernel regression (KR) combines the Euclidean metric with Gaussian kernels~\cite{regression}. Using Euclidean as its learning metric, KR is popular for its simplicity. However, there exist two important drawbacks in KR. Firstly, KR adopts Euclidean metric on the input space, which may preclude its usage on some data sets. For example, if a data set whose features represent completely disparate quantities, the Euclidean metric is meaningless to it.
Secondly, Euclidean metric assigns the same weight $1$ to all the features. It's known that on most of data sets, some features are possibly irrelevant to the regression task. These features ideally should not contribute to this distance metric at all. Therefore, learning a proper distance metric for kernel regression becomes an important issue.

As a well-established nonlinear regression method being widely used in statistics, kernel regression has attracted much attention from researchers. In 2006, Takeda et al. proposed a novel kernel regression algorithm for image denoising\cite{Takeda}. They learn a Mahalanobis matrix from statistics of the local pixel space. However, their algorithm is restricted to specific applications in some sense and cannot be generalized to all cases. At the same year, Keller et al. applied neighborhood component regression to function approximation for reinforcement learning\cite{Keller}. Among all the algorithms related to kernel regression, the most popular one is proposed by Weinberger et al., which they refer to as \emph{Metric Learning for Kernel Regression} (MLKR)\cite{4}. Their work can be interpreted as learning a Mahalanobis matrix for a Gaussian regression kernel via minimization of loss function. However, MLKR just learns a Mahalanobis metric for kernel regression, which does not involve sparse metric learning.  Currently, sparse metric learning is a hot issue in machine learning and data mining. Its objective is to learn a sparse metric which is capable of removing redundancy or noise in data and improving the performance of supervised or
unsupervised learning algorithms\cite{5,26,15,17}.  Over the past several years, many sparse metric learning algorithms have been proposed, including sparse metric learning via linear programming,  sparse metric learning via smooth optimization, and so on\cite{11,16,17}. Unluckily, all the currently published algorithms on sparse metric learning are about classification. Classification refers to an algorithmic procedure for assigning a given piece of input data into one of a given number of categories. The number of categories is finite and the categories are discrete. As to regression, it deals with approximating the probability distribution underlying the data and finding out the most precise target values for the input data. Currently, few sparse metric learning methods are proposed basing on kernel regression issues.

Inspired by the latest innovations on sparse metric learning for classification\cite{16}, we propose a novel kernel regression method to learn a sparse metric over the input space. This metric gives rise to an appropriate kernel function with parameters determined completely from the data. In addition to performing regression, our algorithm can also remove redundancy or noise in data leading to dimensionality reduction. We target the objective of sparse metric learning  directly by enforcing a mixed $(2,1)$-norm regularization over the metric matrix. A Mahalanobis metric is learnt by minimizing the loss function and the metric matrix's mixed $(2,1)$-norm regularization. Experiments on $19$ data sets for regression are performed. The proposed method is also applied to forecast short-term traffic flows to verify its effectiveness. Because KR$\_$SML has the capability of dimensionality reduction, we therefore add another KR model into the comparison. The KR model is referred to as KR$\_$PCA, which first conducts dimensionality reduction on the regressors via principal component analysis (PCA)~\cite{PCA} and then runs KR on the leading principal components. Comparisons with three related kernel regression algorithms under two criterions reveal the competitiveness of the proposed method. In addition, our work is the first to combine kernel regression with sparse metric learning.

A preliminary study of the proposed kernel regression method applied to low-dimensional traffic flow forecasting was presented at a conference\cite{rqhuang}. In this paper, we introduce the detailed framework, provide theoretical justifications, and evaluate the proposed method on more standard regression data sets as well as high-dimensional traffic flow forecasting. The paper is organized as follows. In the next section, some related works about kernel regression are briefly reviewed. The notations establishment and kernel regression problem settings are also presented in this section. In Section~3, we thoroughly introduce the proposed method, namely kernel regression with sparse metric learning(KR$\_$SML). Section~4 reports the experimental results on 19 benchmark data sets for regression, including comparisons with other three related kernel regression methods. In Section~5, the proposed method is applied to forecasting short-time traffic flows. Finally, conclusions and future work are presented in Section~6.

\vspace{11pt}
\section{Kernel Regression and Distance Metric Learning}
Our work integrates kernel regression and sparse metric learning. In this section, we will give a brief review on related works including the traditional kernel regression algorithm with the Euclidean metric (KR), Mahalanobis metric learning for kernel regression (MLKR) and latest developments in distance metric learning. In the following subsections, some notations used in this paper will be introduced before presenting these specific techniques.

\subsection{Basic Notations}
Let $(\vec{x},~y)$ represent an example with input $\vec{x}=(x_1, x_2, \ldots, x_d) \in R^{d}$ and its corresponding target value $y$. In regression case, the target value $y \in R$ and y is continuous. As for a classification issue, $y$ is the label information of an example and it belongs to an integer set (to simplify the classification task, we adopt an integer set to represent the label information). A data set with $n$ examples is denoted by $Z=\{{(\vec{x}}_{i}, y_{i})_{i=1}^n\}$. The space of symmetric $d$ by $d$ matrices is denoted by $S^{d}$. If $S \in S^{d}$ is positive semi-definite, we write it as $S\geq 0$. The cone of positive semi-definite matrices is denoted by $S_{+}^d$ and we denote the set of $d$ by $d$ orthonormal matrices by ${\rm O}^d$. The trace operation for matrices is denoted by $Tr(\cdot)$, which is the sum of all the diagonal elements of a matrix\cite{18}. For any matrix $X,Y$~$\in R^{n\times d},  \langle X,Y\rangle:=Tr(X^TY)$. In addition, any $d$ by $d$ diagonal matrix is denoted by $diag(D_{11},D_{22},\ldots, D_{dd})$, where $D_{11},D_{22},\ldots, D_{dd}$ are the diagonal elements of the matrix.

A Mahalanobis metric is a generalization of the Euclidean metric, in which the squared distance between two examples ${\vec{x}}_{i}$ and ${\vec{x}}_{j}$ is defined as
\begin{equation}
d({\vec{x}}_{i},{\vec{x}}_j)={\left\| {{\vec{x}}_{i} - {{\vec{x}}_j}}
\right\|}^2=({\vec{x}}_{i}-{\vec{x}}_{j})^TM({\vec{x}}_{i}-{\vec{x}}_{j})
\\,
\end{equation}
where $M$ can be any symmetric positive semi-definite real matrix. Setting $M$ as the identity matrix, it recovers the standard Euclidean metric.
Actually, the Mahalanobis distance can be expressed as the Euclidean distance after a mapping ${\vec{x}}\rightarrow A{\vec{x}}$:
\begin{equation}
d({\vec{x}}_{i},{\vec{x}}_j)={\left\|A( {{\vec{x}}_{i} - {{\vec{x}}_j}})
\right\|}^2=({\vec{x}}_{i}-{\vec{x}}_{j})^TA^TA({\vec{x}}_{i}-{\vec{x}}_{j})
\\
\end{equation}
Therefore, $M$ can be decomposed as
\begin{equation}
M=A^TA
\\,
\end{equation}
where $A$ is the decomposition of matrix $M$.
For simplification, we denote the difference of two vectors by ${\vec{x}}_{ij}={\vec{x}}_i-{\vec{x}}_j$.

\subsection{Kernel Regression}

Kernel regression is an estimation technique for fitting data~\cite{regression}. Standard kernel regression usually combines the Gaussian kernel function and Euclidean metric. Its task is to estimate an unknown function $f:R^d \rightarrow R$ based merely on a training set of possibly noisy examples. In kernel regression, the target value of every test input is estimated as
\begin{equation}
y_i=f({\vec{x}}_i)+\varepsilon\\,
\end{equation}
where $\varepsilon $ represents some small noise. \\

Nadaraya and Watson proposed to estimate $f$ as a locally weighted average, using a kernel as the weighting function\cite{2,3}.  Therefore, the estimated value $\hat y_i  \approx f({\vec{x}}_i)$  is approximated by
\begin{equation}
\hat y_i = \frac{{\sum\limits_{j = 1}^k {y_j K_{ij} }
}}{{\sum\limits_{j = 1}^k {K_{ij} } }}\\,
\end{equation}
where $k$ is the number of the test input ${\vec{x}}_i$'s nearest neighbors, $K_{ij}$ is the value of the kernel function based on the distance
between ${\vec{x}}_i$ and its corresponding nearest neighbor ${\vec{x}}_j$.\\\\

The kernel function $K_{ij}=K(d({\vec{x}}_i,{\vec{x}}_j))$ is nonnegative. It is formulated as follows

\begin{equation}
K({\vec{x}}_i,{\vec{x}}_j)= \frac{1}{{\sigma \sqrt {2\pi }}}e^{ -
\frac{{d({\vec{x}}_i ,{\vec{x}}_j )}}{{2\sigma ^2 }}} \\,
\end{equation}
where $d({\vec{x}}_i,{\vec{x}}_j)$ is the Mahalanobis distance
between ${\vec{x}}_i$ and ${\vec{x}}_j$ with $M$ being the metric matrix. Setting $M$ to be the identity matrix $I$, it refers to the Euclidean metric.

The quadratic error loss function of kernel regression is generally formulated as $L$:
\begin{equation}
 L=\sum\limits_{i = 1}^n {(y_i  - \hat y_i )^2 }  \\,
\end{equation}
where $\hat y_i$ represents the estimator of $y_i$ and is computed by formula (5). $L$ represents the accumulated quadratic leave-one-out regression error
of all test examples.

Metric learning for kernel regression (MLKR) aims at training a Mahalanobis matrix by minimizing the error loss of all training examples:
\begin{equation}
\textbf{L}= \sum\limits_{i = 1}^{N_{training}} {(y_i  - \hat y_i )^2 }  \\.
\end{equation}
A transformation matrix $A$, the decomposition of matrix $M$ is learnt by a gradient descent procedure:
\begin{equation}
\Delta A=-\alpha \frac{{\partial \textbf{L}}}{{\partial A}}  \\,
\end{equation}
where $\alpha$ is an adaptive step-size parameter.
Then the Mahalanobis matrix is obtained by $M=A^TA$, and it is used to estimate the targets of test inputs.

\subsection{Distance Metric Learning}

Metric learning is an important research area in machine learning and data mining. A metric is a distance function on a set of points,
mapping pairs of points into the nonnegative real numbers. The objective of metric learning is to find a proper distance function
to make the distance between similar examples as small as possible while enlarging the distance between dissimilar examples. A proper
metric $d(\cdot, \cdot)$ obeys four properties\cite{5}:

\begin{itemize}

\vspace{-6pt}
\item  Nonnegativity: $d({\vec{x}}_i,{\vec{x}}_j)\geq 0$,

\vspace{-6pt}
\item  Symmetry:  $d({\vec{x}}_i, {\vec{x}}_j)= d({\vec{x}}_j, {\vec{x}}_i)$,

\vspace{-6pt}
\item  Triangle inequality:  $d({\vec{x}}_i, {\vec{x}}_j)+ d({\vec{x}}_i, {\vec{x}}_k)\geq d({\vec{x}}_j, {\vec{x}}_k)$,

\vspace{-6pt}
\item  Distinguishability: $d({\vec{x}}_i, {\vec{x}}_j)=0 \Leftrightarrow {\vec{x}}_i = {\vec{x}}_j$.

\vspace{-6pt}
\end{itemize}

For many machine learning algorithms, the choice of a distance metric has a critical influence on their performance. Take the
\emph{k}NN algorithm as an example\cite{33,16}, most implementations of \emph{k}NN adopt the simple Euclidean as their distance metrics.
As mentioned in the previous section, Euclidean metric has two important drawbacks, which badly prevents the algorithms from performing well.
Furthermore, it has been revealed that even a simple linear transformation of the input features can lead to significant improvements in \emph{k}NN classification\cite{5}. Distance metric learning has attracted much attention from researchers and a number of improved algorithms have been proposed.
It is already demonstrated that \emph{k}NN algorithms can be greatly improved by learning an appropriate distance metric from labeled examples\cite{7,8,26}.

A good distance metric should generally preserve the proximity relationship of the data in the transformed space. That is, the distance between similar examples should be relatively smaller than that between dissimilar examples in the transformed space. In supervised classification, the label information can tell us whether two examples are in the same class (similar) or in different classes (dissimilar). For semi-supervised clustering, the side information conveys the information that a couple of examples are similar or dissimilar to each other \cite{6,17,16}. Throughout most of the published works on metric learning for classification, they have two points in common.

\begin{itemize}
\vspace{-6pt}
\item Firstly, triplets containing 3 examples with two kinds of labels are constructed. That is,\\

$T=\{\tau=(i, j, k):i,j, k\in N,\forall({\vec{x}}_i,{\vec{x}}_j)\in S~$\mbox{and}$~({\vec{x}}_j,{\vec{x}}_k)\in D\}$,\\\\ 
where $S$ represents pairs of similar examples and $D$ represents the dissimilar pairs according to the label information.

\vspace{-6pt}
\item Secondly, the distances between examples should always satisfy some constraints: the distance of similar examples are smaller than that of dissimilar examples.

 $ {\|A({\vec{x}}_j-{\vec{x}}_k)\|}^2\geq {\|A({\vec{x}}_i-{\vec{x}}_j)\|}^2+1, \forall (i,j,k)\in T.$\\

\vspace{-6pt}
\end{itemize}

Algorithms of metric learning for classification have strongest dependence on the label information. Among the existing distance metric learning algorithms, distance metric learning for large margin nearest neighbor classification (LMNN)\cite{5} is one of the most prominent. LMNN is motivated from the large margin concept. It learns a Mahalanobis metric with the goal that the \emph{k}-nearest neighbors of the test input always belong to the same class while examples from different classes are separated by a large margin.

Distance metric learning usually attempts to learn a distance function $f$ with a full-rank square metric matrix $M$ from the given data set. However, the observed data are probably contaminated by noise or redundancy,  especially for high-dimensional data sets. If the contaminated data are not pre-processed to remove the irrelevant information in data, the accuracy of related algorithms may be degraded to an extent. Learning a full-rank metric matrix can not suppress the noise and will instead make the learning process time-consuming and expensive. To overcome the drawbacks of existing distance metric learning algorithms, a number of sparse metric learning approaches have been proposed\cite{11,16,17}. In particular, the latest innovation of metric learning proposes a unified framework for sparse metric learning (GSML)\cite{16}. Existing sparse metric learning algorithms are able to learn a good distance metric as well as a sparse or low-dimensional representation. Specifically, GSML directly targets the shortcomings of existing distance metric learning algorithms and has been demonstrated considerable improvement.\\

GSML bases itself on two principal hypotheses:
\begin{itemize}
\vspace{-6pt}
\item  the distance between similar examples should be relatively smaller than that of dissimilar examples;

\vspace{-6pt}
\item a good distance metric should have the capability of removing noise in data leading to dimensionality reduction.

\vspace{-6pt}
\end{itemize}

Therefore, to meet the first hypothesis, the distances of a triplet in the transformed space should satisfy a constraint condition:
\begin{equation}
{\|A({\vec{x}}_j-{\vec{x}}_k)\|}^2\geq {\|A({\vec{x}}_i-{\vec{x}}_j)\|}^2+1,\\
\forall (i,j,k)\in T,\\
\end{equation}
where $A\in R^{d\times d}$ is a transformation matrix. For the second hypothesis, any transformation vector $\hat{\vec{x}}_i=A{\vec{x}}_i$
should have fewer dimensions than the input vector ${\vec{x}}_i$. Let $A_i$ denote the $i$-th row vector of $A$, if $\| A_i\|$=0, then
the $i$-th entry of ${\vec{x}}_i$ becomes 0. Thus, to get a sparse solution, we can enforce a $L_1$-norm regularization across the vector
$(\| A_1\|,\| A_2\|,\ldots,\| A_d\|)$, i.e., $\sum\limits_{i = 1}^d {\left\| {A_i } \right\|} $. That is, the sparse representation is realized
by enforcing a mixed $(2,1)$-norm regularization over the transformation matrix $A$. Ideally, the principal components of $\hat {\vec{x}}_i$ are
expected to be sparse. Therefore, an extra orthonormal transformation matrix $U\in O^d$ is introduced and $\hat{\vec{x}}_i=AU{\vec{x}}_i$.

To avoid the situation that there is no solution for equation~(10), slack variables ${\xi}$ are introduced.
After a serials of transformation, the sparse metric learning formulation is proposed\cite{16,17},\\
\begin{equation}
\begin{array}{l}
\mathop {\min }\limits_{U \in O^d } \mathop {\min }\limits_{M \in \mathop S\nolimits_ + ^d } \sum\nolimits_\tau  {\xi _\tau  }  + \gamma \left\| M \right\|_{(2,1)}  \\
 s.t. ~~~1 + {\vec{x}}_{ij}^T U^T MU{\vec{x}}_{ij}  \le {\vec{x}}_{jk}^T U^T MU{\vec{x}}_{jk}  + \xi
 _\tau, \\
 ~~~~~~~~\xi _\tau   \ge 0,\forall \tau  = (i,j,k) \in T.\\
\end{array}
\end{equation}
With reference to \cite{17,18,30,22}, problem (11) is equivalent to the following convex optimization problem:
\begin{equation}
\begin{array}{l}
\mathop {\min }\limits_{M \in \mathop S\nolimits_ + ^d } \sum\nolimits_\tau  {\xi _\tau  }  + \gamma Tr(M)  \\
 s.t. ~~~1 + {\vec{x}}_{ij}^TM{\vec{x}}_{ij}  \le {\vec{x}}_{jk}^TM{\vec{x}}_{jk}  + \xi
 _\tau, \\
 ~~~~~~~~\xi _\tau   \ge 0,\forall \tau  = (i,j,k) \in T.\\
\end{array}
\end{equation}
By defining the hinge loss $[z]_{+}=\left\{ \begin{array}{l}
 z ~~~\mbox{if}~ z>0\\
 0 ~~~\mbox{otherwise}\\
 \end{array} \right.
$, the above problem can be transformed into an unconstrained optimization problem:
\begin{equation}
\mathop {\min }\limits_{M\geq 0 }\sum\nolimits_\tau  {{[1 +
{\vec{x}}_{ij}^TM{\vec{x}}_{ij}  -
{\vec{x}}_{jk}^TM{\vec{x}}_{jk}]}_{+}  } + \gamma Tr(LM)
\\.
\end{equation}
Then the matrix $M$ is obtained by a standard sub-gradient descent procedure.

\vspace{12pt}
\section{Kernel Regression with Sparse Metric Learning (KR$\_$SML) }

We aim at learning a low-rank Mahalanobis matrix for kernel regression. Besides preserving the proximity relationship of examples in the transformed space, a good distance metric should be able to effectively remove possible noise in data leading to dimensionality reduction. Therefore, for the purpose of noise suppression, the metric matrix defined by $M$ should be regularized. We will give a detailed explanation on how the metric matrix can be regularized for forcing sparsity. In the end, we introduce the sparse regularization of the metric matrix into the quadratic error loss function of kernel regression to build a sparse metric learning model for kernel regression.

As mentioned in previous sections, existing distance metric learning algorithms usually attempt to learn a distance function $f$ with a full-rank square metric matrix $M$ from the given data set. However, the observed data are probably contaminated by noise, especially for high-dimensional data sets. The noise in the data may prevent the algorithms from performing well if the contaminated data are not pre-processed to remove the noise. Worse still, learning a full-rank metric matrix can not suppress the noise but instead will make the learning process time-consuming and expensive. Therefore, to conquer the drawbacks of existing distance metric learning algorithms, sparse metric learning emerges as required. However, all the existing sparse metric learning algorithms are about classification, none for regression. Therefore, to learn a sparse metric for kernel regression becomes meaningful and necessary.

The Mahalanobis metric is learnt in the transformed space, after the mapping:
\begin{center}
${\vec{x}}\rightarrow A{\vec{x}}$,
\end{center}
where $A\in R^{d\times d}$ is the transformation matrix and the metric matrix $M$ can be obtained by $M=A^TA$. With the purpose that a good distance metric should be able to remove noise in data leading to dimensionality reduction. Ideally, the principal components of input vector ${\vec{x}}_i$ are expected to be sparse. Therefore, any transformation vector $\hat {\vec{x}}_i$ ($\hat{\vec{x}}_i=A{\vec{x}}_i$ )  should have fewer dimensions than its corresponding
input vector ${\vec{x}}_i$. As a $1$-norm regularization over a vector can produce a sparse vector, certain columns of $A$ should be expected to become zero vectors. That is the basic motivation of sparse metric learning\cite{17}.

Let $A_i$ denote the $i$-th row vector of $A$, if $\| A_i\|$=0, then the $i$-th entry of ${\vec{x}}_i$ becomes 0. Thus, enforcing a $L_1$-norm regularization over the vector $(\| A_1\|,\| A_2\|,\ldots,\| A_d\|)$, i.e., $\sum\limits_{i = 1}^d {\left\| {A_i } \right\|} $, leads to a sparse representation of $A$. Thereinto, $\sum\limits_{i = 1}^d {\left\| {A_i } \right\|} $ represents the mixed $(2,1)$-norm regularization of matrix $A$. Therefore, the sparse representation can be realized by enforcing a  mixed $(2,1)$-norm  regularization over the transformation matrix $A$. On the other hand, the metric matrix $M=A^TA=(M_1, M_2,\ldots,M_d)$. It is obvious that $M_i\equiv 0$ is equivalent to $A_i\equiv0$. Motivated by this observation, instead of enforcing a $L_1$-norm regularization over the vector $(\|A_1\|,\|A_2\|,\ldots,\|A_d\|)$, $L_1$-norm regularization can be enforced across the vector
$(\| M_1\|,\| M_2\|,\ldots,\| M_d\|)$. The $(2,1)$-norm regularization over $M$ is denoted by $\| M\|_{(2,1)}=\sum\limits_{i = 1}^d {\left\| {M_i } \right\|}$.
A similar mixed $(2, 1)$-norm regularization is used for multi-task learning and multi-class classification to learn the sparse representation shared across different tasks or classes~\cite{8,22}.

The task of kernel regression is to estimate the target values for test inputs as precise as possible. In other words, the goal of kernel regression is to make the accumulated quadratic regression error, the concrete form of which is $L=\sum\limits_{i = 1}^n {(y_i  - \hat y_i )^2 }$, as small as possible. Generally speaking, if a kernel regression algorithm with a specific distance metric performs well on the training data, it often gets a good performance on the test data as well. Therefore, following the idea of MLKR~\cite{5}, we train a Mahalanobis metric matrix by the training data. Bearing the objective of kernel regression and sparse metric learning in mind, our work would ensure that the error is small and the metric matrix is sparse. Therefore, a Mahalanobis metric matrix is learnt by making the loss function of kernel regression and the  mixed  $(2,1)$-norm regularization over $M$ to a minimum. That is, the objective function of the proposed kernel regression algorithm (KR$\_$SML) is the minimum of $L(M)$, which is represented as follows\\
\begin{equation}
L(M)=\sum\limits_{i = 1}^{N_{training}} {(y_i  - \hat y_i )^2 }+\mu
\|M\|_{(2,1)}\\.
\end{equation}\\

The mixed $(2,1)$-norm regularization over $M$ in the objective
function is non-convex and non-differentiable. Actually, the minimum of the mixed $(2,1)$-norm regularization over $M$ is equivalent to
the trace of $M$\cite{17}, which is presented as Theorem $1$.\\

\begin{theorem}

$\emph{min} \|M\|_{(2,1)}=Tr(M)$\\\\
\textbf{Proof.} By the eigen-decomposition of $M$ there exists ~$V\in O^d$ such that $M=V^T\lambda(M)V$. Thereinto, the
diagonal matrix $\lambda(M)=diag({\lambda}_1,{\lambda}_2,\ldots,{\lambda}_d)$, where ${\lambda}_i$
is the $i$-th eigen-value of $M$. Therefore, $\|M\|_{(2,1)}={\|V^T\lambda(M)V\|}_{(2,1)}$. Observing that

\begin{equation}
\begin{array}{l}
 \|V^T \lambda (M)V\|_{(2,1)}  = \sum\nolimits_i {(\sum\nolimits_j {(\sum\nolimits_k {V_{ki} \lambda _k V_{kj} } )^2 } )^{\frac{1}{2}} }  \\
  = \sum\nolimits_i {(\sum\nolimits_{k,k'} {(\sum\nolimits_j {V_{ki} V_{k'i} } )\lambda _k V_{kj} \lambda _{k'} V_{k'j} } )^{\frac{1}{2}}
  }\\
  = \sum\nolimits_i {(\sum\nolimits_k {\lambda _k ^2 } V_{ki} ^2 )^{\frac{1}{2}}
  },\\
 \end{array}
\end{equation}
in the last equality, we use the fact that $V\in O^d$, i.e.,
${\sum\nolimits_j {V_{kj} V_{k^{'}j} } }={\delta}_{kk^{'}}$.
Applying Cauchy-Schwartz's inequality implies that $\sum\nolimits_k
{\lambda _k V_{ki} ^2 }  \le (\sum\nolimits_k {\lambda _k ^2 V_{ki}
^2 } )^{\frac{1}{2}} (\sum\nolimits_k {V_{ki} ^2 } )^{\frac{1}{2}} =
(\sum\nolimits_k {\lambda _k ^2 V_{ki} ^2 } )^{\frac{1}{2}} $.
Putting this back into equation (15) and the following result is obtained
\begin{equation}
\|M\|_{(2,1)}  \ge \sum\nolimits_i {\sum\nolimits_k {\lambda _k
V_{ki} ^2 } }  = \sum\nolimits_k {\lambda _k }  = Tr(M)\\.
\end{equation}
If we make $V$ the identity matrix $I$, then the minimum of
$\sum\nolimits_i {(\sum\nolimits_k {\lambda _k ^2 } V_{ki} ^2
)^{\frac{1}{2}} } $ is equal to $Tr(M)$. That is

\begin{equation}
\min \|M\|_{(2,1)}= Tr(M)\\.
\end{equation}

\end{theorem}

Therefore, the objective function (14) is equivalent to

\begin{equation}
L(M)=\sum\limits_{i = 1}^{N_{training}} {(y_i  - \hat y_i )^2 }+ \mu Tr(M)\\.
\end{equation}

With reference to \cite{4,12} and making use of $\frac{{\partial Tr(M)}}{{\partial
M}} = I$, the gradient of (18) with respect to $M$ can be stated as
\begin{equation}
\frac{{\partial L(M)}}{{\partial M}}{\rm{ = 2}}\sum\limits_{i = 1}^{N_{training}}
{(\hat y_i  - y_i )\frac{{\sum\limits_{j = 1}^k {(\hat y_i  - y_j
)K_{ij} {\vec{x}}_{ij} {\vec{x}}_{ij} ^T } }}{{\sum\limits_{j = 1}^k
{K_{ij} } }}} + \mu I \\.
\end{equation}

After setting the initial value of $M$, we adjust its subsequent values
using a gradient descent procedure. In each iteration, we have
to keep $M$ positive semi-definite. Let $G^t$ denote the gradient of
the objective function at the $t$-th iteration, then $G^t$ can be
stated as
\begin{equation}
G^t{\rm{ = 2}}\sum\limits_{i = 1}^{N_{training}} {(\hat y_i  - y_i
)\frac{{\sum\limits_{j = 1}^k {(\hat y_i  - y_j )K_{ij}
{\vec{x}}_{ij} {\vec{x}}_{ij} ^T } }}{{\sum\limits_{j = 1}^k {K_{ij}
} }}} + \mu I \\.
\end{equation}
At each step, the metric matrix $M$ can be updated by
\begin{equation}
M_{(t)}  = M_{(t - 1)}  - \alpha G^t\\,
\end{equation}
where $\alpha$ is a small positive step-size constant. We then
project the matrix $M_{(t)}$ to the cone of positive semi-definite
matrices by the eigen-decomposition of matrix $M_{(t)}$, i.e.,
$M_{(t)}=P^T\Lambda P$, where $P$ is the eigen-vector matrix, and
$\Lambda$ is the diagonal matrix with the diagonal elements
${\lambda}_i$ being the eigen-values of $M_{(t)}$. To keep $M_{(t)}$
semi-definite, we set $M_{(t)}=P^T{\Lambda}_{+} P$,
where
${\Lambda}_{+}=diag(max\{0,{\lambda}_1\},max\{0,{\lambda}_2\},\ldots,max\{0,{\lambda}_d\})$.

According to the above details, the proposed algorithm is illustrated as
follows.
\begin{itemize}
\item \textbf{Begin}
\item \textbf{Input} Matrix $M$,~step-size $\alpha$ for adapting $M$, step-size $\mu$ for adapting $L(M)$, stop criterion $\theta$,~$t \leftarrow
0$.
\begin{itemize}
\item \textbf{Do} $t \leftarrow t+1$
\item Compute the gradient of objective function $G^t$ at the $t$-th iteration.
\item $M_{(t)}  \leftarrow M_{(t - 1)}  - \alpha G^t$
\item $M_{(t)}\leftarrow P^T{\Lambda}_{+} P$
\item Compute the value of function $L(M_t)$ at the $t$-th iteration.
\item \textbf{Until} $|L(M_t)-L(M_{t-1})| \leq{\theta}$.
\end{itemize}
\item \textbf{Output} $M$
\item \textbf{End}
\end{itemize}

\vspace{12pt}
\section{Experiments}

The objective of the proposed kernel regression algorithm is to learn a good distance metric for kernel regression and simultaneously remove the noise of data leading to dimensionality reduction. In our experiments, we compare the proposed algorithm with three other competing kernel regression algorithms KR, MLKR, and KR$\_$PCA on $19$ standard regression data sets. Information about the data sets, experimental settings and results will be presented in this section in detail.

\subsection{Data Description and Configuration}
The first $16$ standard regression data sets utilized in our experiments are
from \emph{Data for Evaluating Learning in Valid Experiments
(Delve)}\footnote{\emph{Delve}: \url{http://www.cs.toronto.edu/~delve/data/datasets.html}}. They are Kin family of data sets and Pumadyn family of data sets \footnote{More information about the specific data set is available at \url{http://www.cs.toronto.edu/~delve/data/kin/desc.html} and
\url{http://www.cs.toronto.edu/~delve/data/pumadyn/desc.html}}. They are generated by two synthetic robot arms. Half of the sixteen data sets have $32$ dimensions and the other half are of dimension $8$. Each data set has already been randomly split into four disjoint training sets of size $n=1024$ and four corresponding test sets of the same size. As a result, we get four training sets and four test sets. The final result is the mean of the results of the four individual runs. There are other papers that have used the DELVE data sets in assessing model performance~\cite{RN,GP}. The last $3$ data sets are from UCI machine learning repository: \url{http://archive.ics.uci.edu/ml/} for regression. For the last $3$ data sets, the final results are given as an average over $10$ random splits of the data. The detailed information about the $19$ data sets are listed in Table 1, where \emph{dataset} represents the name of a data set, \emph{n} represents the size of the data set, and \emph{d} represents the number of dimensions.

The  target value of a test input is estimated by a locally weighted average of values of its \emph{k} nearest neighbors in the training set. In our implementation we only considered the $\emph{k}=30$ nearest neighbors of each test examples. The trade-off parameter $\mu$ in the objective function of our
proposed algorithm and the step-size constant $\alpha$ used in KR$\_$SML
and MLKR are tuned by 10-fold cross-validation on training sets. Specifically, KR$\_$SML consists of setting the initial value of metric matrix $M$, and then adjusting its values using a gradient descent procedure. In our experiments, we follow \cite{4} to initialize it with the identity matrix.

\begin{table}[!ht]
\centering \caption{The 19 data sets of regression used for experiments}
\begin{tabular}{llllllll}
\hline\noalign{\smallskip}
dataset&n&d&dataset&n&d& \\
\noalign{\smallskip}\hline\noalign{\smallskip}
 kin8fh     & 8096&8 & puma-8fh & 8096&8&\\
 kin8fm     & 8096&8 & puma-8fm & 8096&8&\\
 kin8nh     & 8096&8 & puma-8nh & 8096&8&\\
 kin8nm     & 8096&8 & puma-8nm & 8096&8&\\
 kin32fh    & 8096&32& puma-32fh& 8096&32&\\
 kin32fm    & 8096&32& puma-32fm& 8096&32&\\
 kin32nh    & 8096&32& puma-32nh& 8096&32&\\
 kin32nm    & 8096&32& puma-32nm& 8096&32&\\
 Concrete   & 1030&8 & housing    & 506 &13&\\
 parkinsons & 5875&21&            &     &  &\\
\noalign{\smallskip}\hline
\end{tabular}
\end{table}%

\subsection{Experimental Results}

The proposed method has the capability of dimensionality reduction. We provide another KR model with principal component analysis (PCA)~\cite{PCA}, one of the most commonly used algorithms for dimensionality reduction, for comparison. The model is referred to as KR$\_$PCA. It first conducts dimensionality reduction on the regressors via principal component analysis and then runs KR on the leading principal components. The other two competing methods are KR and MLKR. To compare the performance of the proposed algorithm with other three related kernel regression algorithms, KR, MLKR, and KR$\_$PCA, two widely-used criterions are adopted to evaluate experimental results. They are root mean squared error ($RMSE$) and mean absolute relative error ($MARE$), respectively,
which are formulated as follows
\begin{equation}
RMSE=\sqrt {\frac{1}{n}\sum\limits_{i = 1}^n {(y_i  - \hat y_i )^2 }}\\,
\end{equation}
and
\begin{equation}
MARE=\frac{1}{n}\sum\limits_{i = 1}^n {\frac{{|y_i  - \hat y_i
|}}{{y_i }}}\\.
\end{equation}
Besides, we also compare the accumulated quadratic leave-one-out regression error of test examples $L=\sum\limits_{i = 1}^n {(y_i - \hat y_i )^2 }$,
where $\hat y_i$ is the estimator of $y_i$ and $n$ is the number of test examples. Moreover, we provide the performance in terms of dimensionality reduction as well.\\

\begin{table}[!ht]
\centering \caption{Experimental results based on $RMSE$ and the final rank of metric matrix $M$ learnt by KR$\_$SML.}
\begin{tabular}{lllllcccc}
\hline\noalign{\smallskip}
 datasets  &KR           &MLKR       &KR$\_$PCA &KR$\_$SML        &d   &Rank($M$)&PCA\_d\\
\noalign{\smallskip}\hline\noalign{\smallskip}
 kin8fh   &0.0511        &0.0485      &0.0511&\textbf{0.0456}     &8   &5        &8\\
 kin8fm   &0.0318        &0.0274      &0.0318&\textbf{0.0181}     &8   &6        &8\\
 kin8nh   &0.1998        &0.1806      &0.1932&\textbf{0.1791}     &8   &6        &8\\
 kin8nm   &0.1616        &0.1206      &0.1440&\textbf{0.1130}     &8   &7        &8\\
 kin32fh  &0.4031        &0.2722      &0.4075&\textbf{0.2671}     &32  &12       &29\\
 kin32fm  &0.3397        &0.1272      &0.3422&\textbf{0.1207}     &32  &4        &29\\
 kin32nh  &0.4904        &0.4897      &0.5585&\textbf{0.4789}     &32  &26       &16\\
 kin32nm  &0.4524        &0.4517      &0.5196&\textbf{0.4326}     &32  &15       &16\\
 puma-8fh &3.4018        &3.3962      &3.3894&\textbf{3.3150}     &8   &2        &6\\
 puma-8fm &1.5902        &1.5502      &1.5791&\textbf{1.1136}     &8   &2        &6\\
 puma-8nh &3.9231        &3.9113      &3.7784&\textbf{3.4174}     &8   &2        &6\\
 puma-8nm &2.8196        &2.8087      &2.4487&\textbf{1.2511}     &8   &2        &6\\
 puma-32fh&0.0214        &0.0210      &0.0293&\textbf{0.0210}     &32  &30       &5\\
 puma-32fm&0.0064        &0.0050      &0.0085&\textbf{0.0050}     &32  &30       &5\\
 puma-32nh&0.0335        &0.0335      &0.0416&\textbf{0.0335}     &32  &32       &5\\
 puma-32nm&0.0273        &0.0273      &0.0344&\textbf{0.0273}     &32  &32       &5\\
 Concrete &8.4746        &8.4624      &8.7907&\textbf{8.1076}     &8   &3        &5\\
 housing  &5.9790        &5.9772      &7.4727&\textbf{5.8919}     &13  &8        &2\\
 parkinsons&0.0555       &0.0555      &0.0739&\textbf{0.0465}     &21  &20       &3\\
 \noalign{\smallskip}\hline
\end{tabular}
\end{table}%

\begin{figure}[thpb]
      \centering
      \includegraphics[scale=1.0]{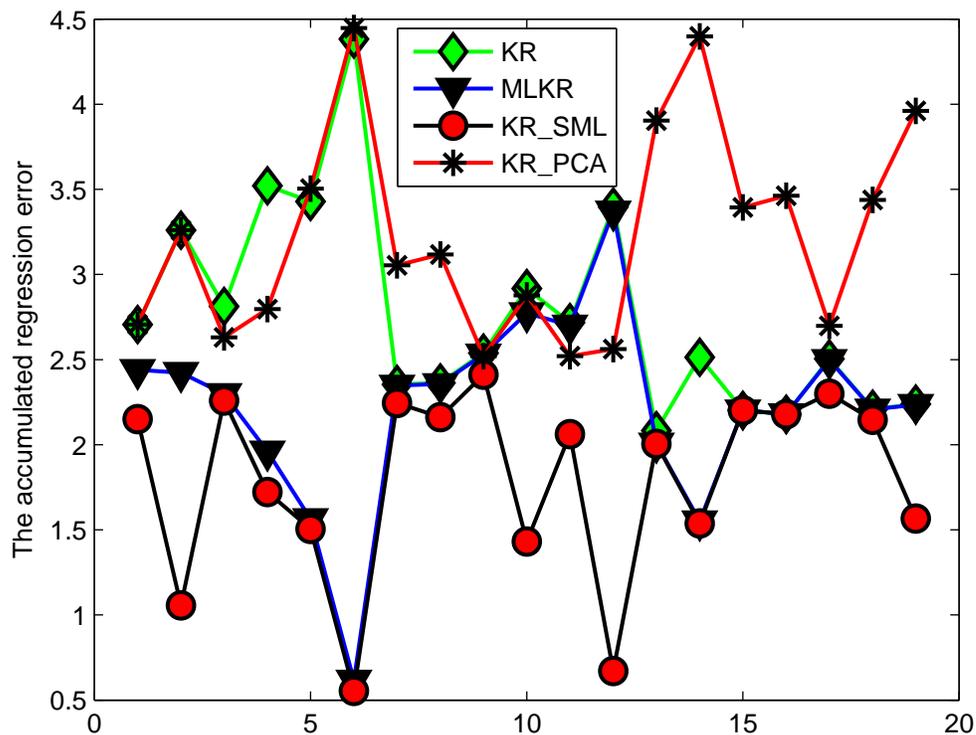}
      \caption{The accumulated regression error of KR, MLKR, KR$\_$PCA and KR$\_$SML on 19 data sets.}
      \label{figurelabel}
\end{figure}

\begin{figure}[thpb]
      \centering
      \includegraphics[scale=1.0]{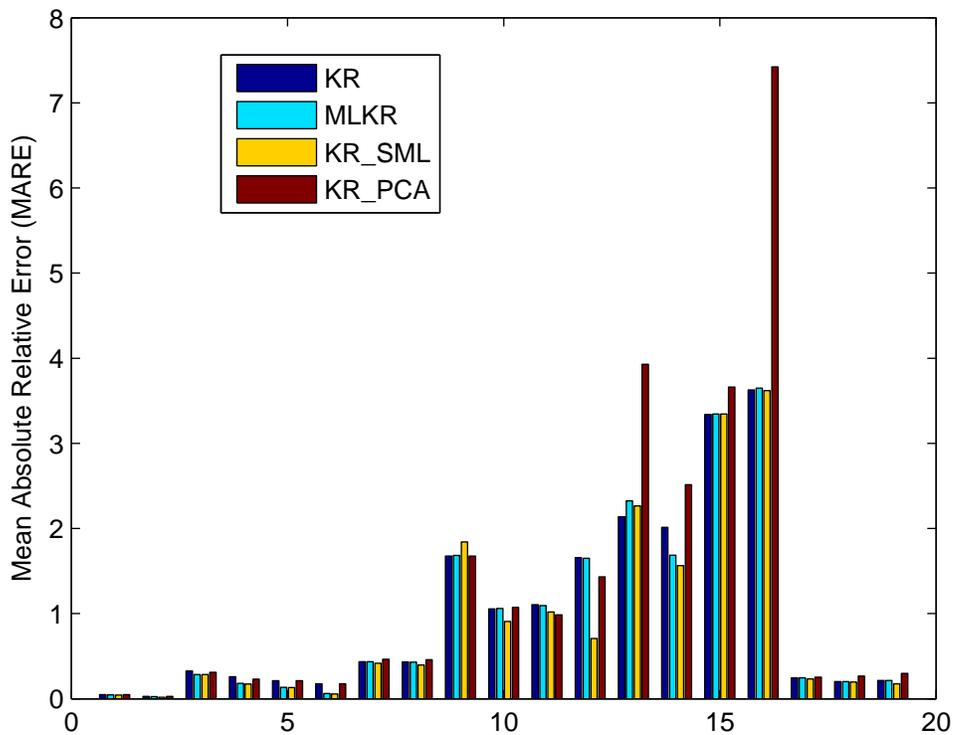}
      \caption{$MARE$ of KR, MLKR, KR$\_$PCA and KR$\_$SML on 19 data sets.}
      \label{figurelabel}
\end{figure}

Experimental results based on criterion $RMSE$ of four kernel regression algorithms on $19$ data sets are shown in Table~2. The bold number in the
table represents that the kernel regression algorithm in the corresponding column performs best on the data set of the corresponding row. The \emph{d} column tells the original dimension of each data set. The Rank($M$) column reports the rank of the metric matrix $M$ learnt by KR$\_$SML.  The PCA\_d column shows the final dimension of each data set learnt by PCA. To give an intuitive comparison, we show the accumulated quadratic leave-one-out regression error of the four kernel regression algorithms in Fig.~1. The final results of four kernel regression algorithms based on comparison criterion $MARE$ are shown in Fig. 2.

\subsection{Discussion}
The objective of kernel regression is to estimate the target values for test inputs as precise as possible to minimize the accumulated quadratic regression error. Therefore, the test error $L$ (or $RMSE$) is treated as the most important comparison  criterion for kernel regression algorithms. As is revealed by the experimental results presented in Table~2 and Fig.~1, the proposed algorithm gets the best performance on all the $19$ data sets. As to the comparison criterion $MARE$, KR$\_$SML also outperforms the other three kernel regression algorithms on most of data sets. In addition, KR$\_$SML is the only kernel regression algorithm that targets the objective of sparse metric learning. According to data in the last three columns of Table~2, the metric matrix $M$ learnt by KR$\_$SML has lower rank than its original dimension on most of data sets. If a sparse or low-rank metric matrix $M$ is learnt, the performance of KR$\_$SML is obviously much better than that of the other three competing kernel regression algorithms. For the data sets puma-32nh and puma-32nm, KR$\_$SML cannot learn a sparse metric matrix $M$. However, all the three kernel regression algorithms  perform well on the two data sets. It can be explained that the features of the two data sets are all meaningful to the distance metric. That is, if the data are contaminated by noise, KR$\_$SML has the capability of learning a sparse metric matrix leading to dimensionality reduction and the performance of KR is improved a lot. On the other hand, if the data are not contaminated by noise, KR$\_$SML cannot learn a low-rank metric matrix, but it can perform as well as the other two kernel regression algorithms. As to KR\_PCA, it also has the ability to reduce dimensionality due to PCA, but it gets the worst performance. This again demonstrates that PCA is easily affected by the scaling of input features and ignores the important information for being totally unsupervised~\cite{4}. When the noise become the leading PCA principal components, it tends to keep the noise and ignore the signal. Therefore, we can conclude that KR$\_$SML can learn a good distance metric and simultaneously remove noise in data leading to dimensionality reduction. As the first work to combine kernel regression with sparse metric learning, KR$\_$SML is a promising and better algorithm for kernel regression.

\vspace{12pt}
\section{KR$\_$SML for Traffic Flow Forecasting}

Short-term traffic flow forecasting is one of the most important and
fundamental problems in intelligent transportation systems (ITS). It
contributes a lot to traffic signal control and congestion
avoidance. The benefits of ITS cannot be realized without the
ability to forecast traffic condition in the next time interval, for
example, 5 minutes to half an hour. A good traffic condition
forecasting model will provide this ability and make traffic
management more efficient. There have been a variety of methods proposed for
short-term traffic flow forecasting such as Markov chain models, time series models, Bayesian networks, support vector
machines and kernel regression. In this paper, the proposed kernel regression algorithm KR$\_$SML
is applied to forecasting short-time traffic flow to evaluate its effectiveness.

The problem addressed in this section is to forecast the future
traffic flow rates at given roadway locations from the historical
data on a transportation network. The data are from Beijing's Traffic Management Bureau. From the real urban
traffic map, we select a representative patch to verify the proposed
approach, which is given in Fig. 3~\cite{Sun}. Each circle
in the sketch map denotes a road junction. An arrow shows the
direction of traffic flow, which reaches the corresponding road link
from its upstream link. Paths without arrows are of no traffic flow
records. Vehicular flow rates of discrete time series are recorded
every 15 minutes. The recording period is 25 days (totally 2400
recorded entries) from March, 2002. In our experiment, the raw data
are divided into two sets, 2112 recorded entries of the first 22
days as the training set and the rest recorded entries as the test
set. For evaluation, experiments are performed with multiple
randomly selected roads from Fig. 3.

Let $x_1$, $x_2$, $x_3$, \ldots, $x_{2400}$ denote the original 2400
ordered recorded data of a road. First the raw data have to be changed into
examples of vector form. An example is represented as $(\vec{x},y)$,
where $\vec{x} \in R^d$ and $y \in R$\cite{12}. In our experiment, the dimension of $\vec{x}$, $d$, and the number of past flows used to forecast the current flow, $k$, are empirically set as $45$ and $8$,
respectively.

\begin{figure}
\centering
\includegraphics[height=7.5cm]{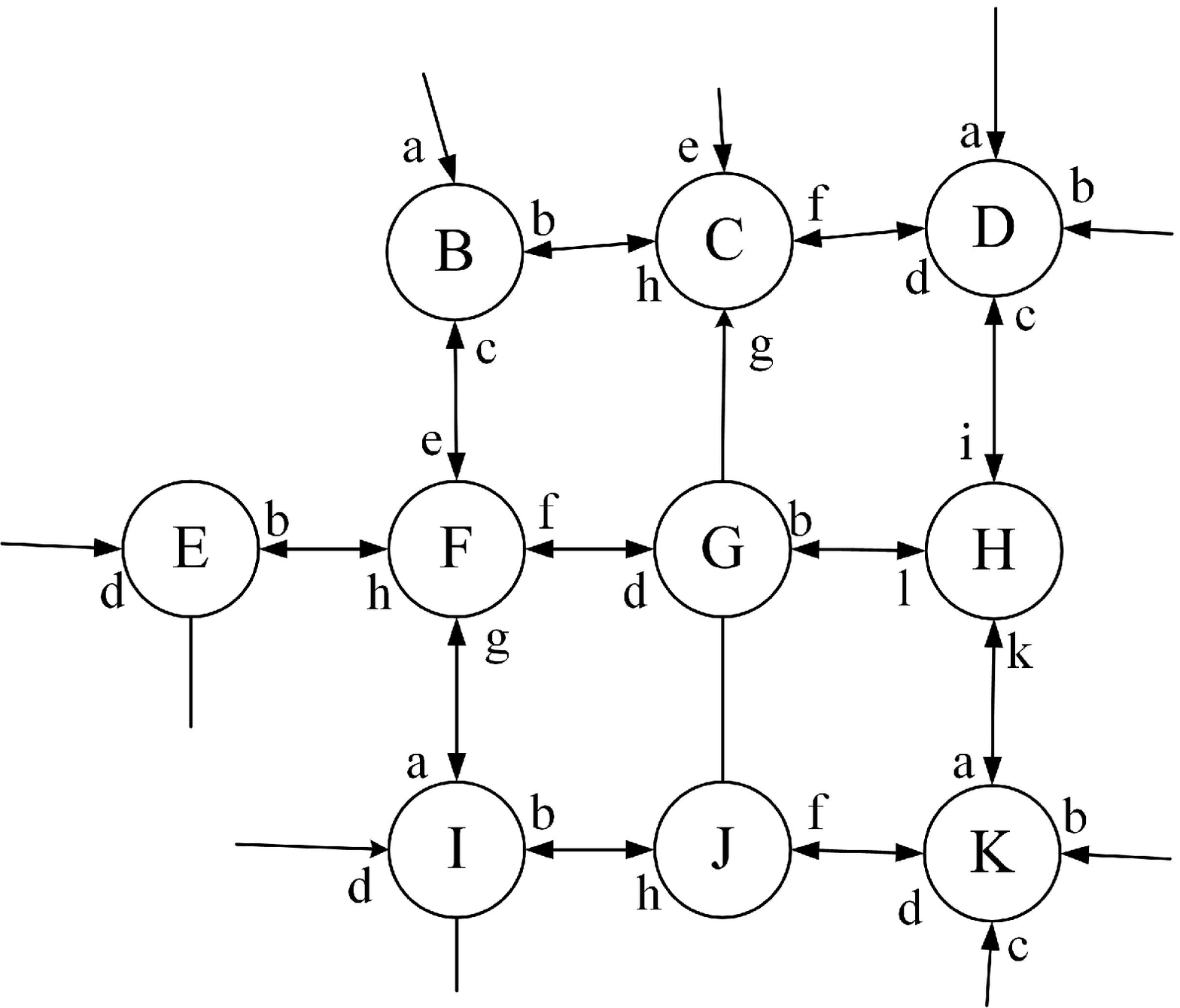}
\caption{A patch of traffic map taken from the East Section of the
Third Circle of Beijing City Map where the UTC/SCOOT system is
mounted.} \label{fig:example}
\end{figure}

\begin{table}[h]
\begin{center}
\caption{The final rank of metric matrix $M$ learnt by KR$\_$SML.}
\begin{tabular}{cccccccccc}

\hline
Road &&d  &&Rank($M$) &PCA\_d\\
\hline
Ba   &&45         &&25&35\\
Cf   &&45         &&35&35\\
Fe   &&45         &&26&34\\
Gb   &&45         &&39&32\\
Hi   &&45         &&41&36\\
\hline
\end{tabular}
\end{center}
\label{t1}
\end{table}

\begin{table}[h]
\begin{center}
\caption{Training error comparison}
\begin{tabular}{lccccccccccccccc}
\hline
\multicolumn{1}{c}{}& \multicolumn{4}{c}{$MARE$}& \multicolumn{4}{c}{$RMSE$}\\
\hline \multicolumn{1}{l}{R}
& \multicolumn{1}{c}{KR}
& \multicolumn{1}{c}{MLKR}
& \multicolumn{1}{c}{KR$\_$PCA}
& \multicolumn{1}{c}{KR$\_$SML}
& \multicolumn{1}{c}{KR}
& \multicolumn{1}{c}{MLKR}
& \multicolumn{1}{c}{KR$\_$PCA}
& \multicolumn{1}{c}{KR$\_$SML}\\
\hline
Ba  &0.229&0.148&0.293&\textbf{0.140} &279.10&177.30&298.54&\textbf{148.61}\\
Cf &0.130&0.118&0.231&\textbf{0.114}&134.58&116.29&198.15&\textbf{106.15}\\
Fe &0.209&\textbf{0.112}&0.265&0.112&341.53&169.00&363.81&\textbf{157.08}\\
Gb&0.235&0.174&0.309&\textbf{0.155}&170.00&135.88&179.37&\textbf{88.15}\\
Hi &0.245&0.168&0.390&\textbf{0.161}&182.26&113.91&194.07&\textbf{92.01}\\
\hline
\end{tabular}
\end{center}
\label{t1}
\end{table}

\begin{table}[h]
\begin{center}
\caption{Test error comparison}
\begin{tabular}{lccccccccccccccc}
\hline
\multicolumn{1}{c}{}& \multicolumn{4}{c}{$MARE$}& \multicolumn{4}{c}{$RMSE$}\\
\hline \multicolumn{1}{l}{R}
& \multicolumn{1}{c}{KR}
& \multicolumn{1}{c}{MLKR}
& \multicolumn{1}{c}{KR$\_$PCA}
& \multicolumn{1}{c}{KR$\_$SML}
& \multicolumn{1}{c}{KR}
& \multicolumn{1}{c}{MLKR}
& \multicolumn{1}{c}{KR$\_$PCA}
& \multicolumn{1}{c}{KR$\_$SML}\\
\hline
Ba  &0.238&0.157&0.302&\textbf{0.152}&318.87&202.92&333.68&\textbf{193.89}\\
Cf &0.115&0.106&0.226&\textbf{0.102}&137.91&115.80&209.99&\textbf{105.15}\\
Fe &0.204&0.117&0.263&\textbf{0.107}&363.12&191.11&386.50&\textbf{166.83}\\
Gb&0.243&0.172&0.284&\textbf{0.152}&178.15&148.13&185.72&\textbf{100.64}\\
Hi &0.242&0.160&0.353&\textbf{0.152}&198.48&119.25&205.46&\textbf{102.97}\\
\hline
\end{tabular}
\end{center}
\label{t1}
\end{table}

\begin{figure}[htb]
\begin{minipage}[b]{0.38\linewidth}
  \centering
  \centerline{\epsfig{figure=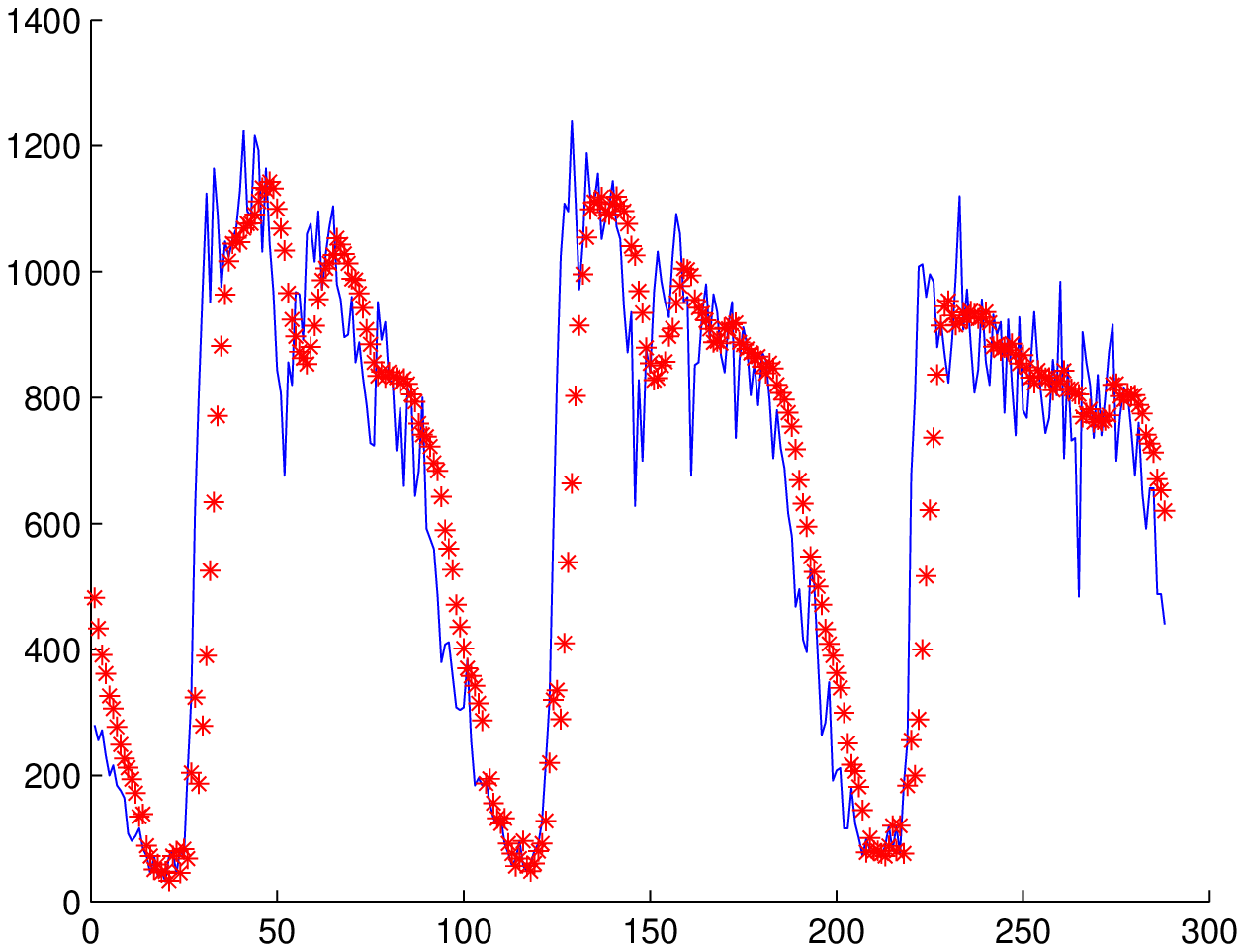,width=6.3cm}}
  \centerline{(a) KR}\medskip
\end{minipage}
\hfill
\begin{minipage}[b]{0.38\linewidth}
  \centering
  \centerline{\epsfig{figure=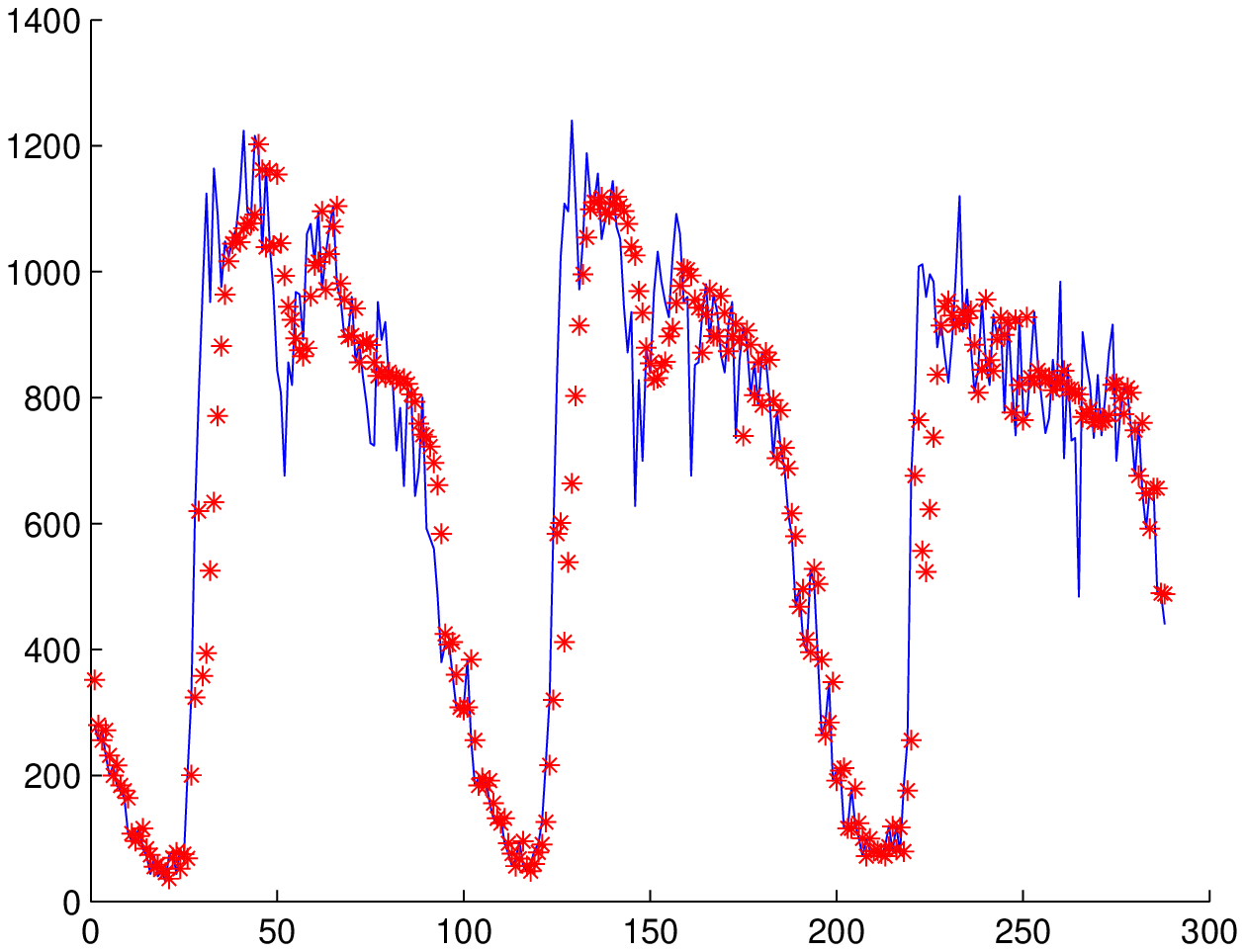,width=6.3cm}}
  \centerline{(b) MLKR }\medskip
\end{minipage}
\hfill
\begin{minipage}[b]{0.38\linewidth}
  \centering
  \centerline{\epsfig{figure=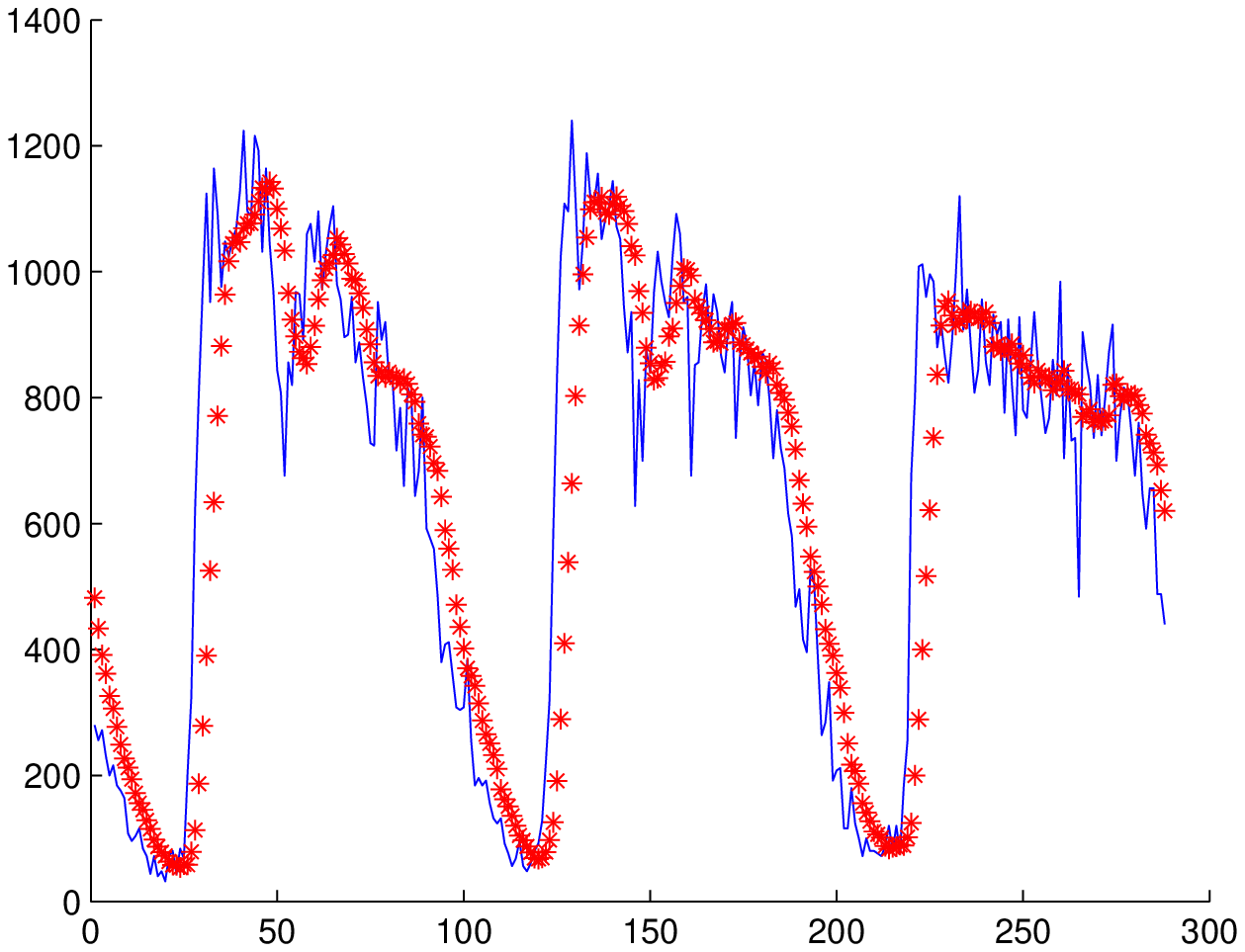,width=6.3cm}}
  \centerline{(c) KR$\_$PCA }\medskip
\end{minipage}
\hfill
\begin{minipage}[b]{0.38\linewidth}
  \centering
  \centerline{\epsfig{figure=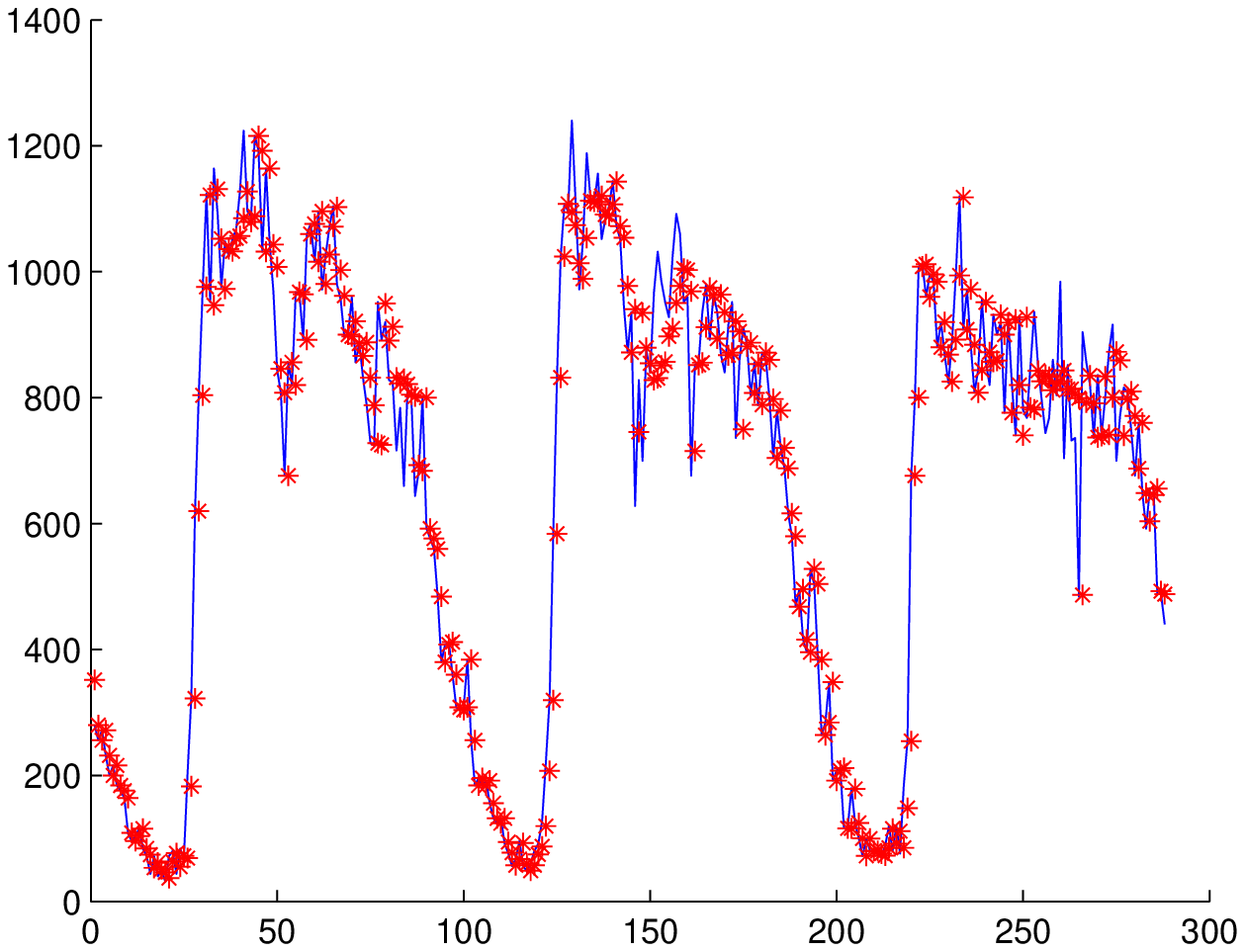,width=6.3cm}}
  \centerline{(d) KR$\_$SML }\medskip
\end{minipage}

\caption{Forecasting results of KR, MLKR, KR$\_$PCA and KR$\_$SML for Gb.}
\label{fig:res}
\end{figure}

The two criterions used to compare the performance of KR$\_$SML, KR$\_$PCA, MLKR and KR in the above section are also adopted in this section.
Detailed information about the rank of the metric matrix $M$ learnt by KR$\_$SML and the final dimension learnt by PCA for
each data set is presented in Table 3. In addition, performance comparison of the four algorithms based on criterions $MARE$ and $RMSE$ on training sets and test sets are reported in Table~4 and Table~5, respectively. In order to give an intuitive illustration
of the forecasting performance, we draw the forecasting results of
Roadway Gb on the test set using KR, MLKR, KR$\_$PCA and KR$\_$SML, which are shown
in Fig. 4, where blue lines represent real recorded data and red
stars represent forecasted results.
Real traffic flow forecasting results reported in Table~3, Table~4
and Table~5 reveal that MLKR and KR$\_$SML are all superior to the
traditional kernel regression algorithm KR and KR\_PCA, which means metric
learning can effectively improve the performance of kernel
regression algorithms. Different from MLKR, the proposed KR$\_$SML is
the first to combine kernel regression with sparse metric learning.
As shown in Table~3, only KR$\_$SML has the capability of learning a
low-rank metric matrix. Furthermore, KR$\_$SML gets much better forecasting results than that of MLKR, KR\_PCA, and KR
on almost all the data sets. Therefore, the conclusion can be drawn that the proposed algorithm is better than KR, KR\_PCA, and MLKR. It can learn
a good metric and effectively remove noise in data leading to dimensionality
reduction as well.

\section{Conclusion}
In this paper, a new kernel regression algorithm with sparse metric learning, which we refer to
as KR$\_$SML is proposed. KR$\_$SML is realized by introducing a mixed $(2,1)$-norm regularization over the metric
matrix $M$ into the objective function of kernel regression. By minimizing the regression error function and the metric matrix's mixed $(2,1)$-norm regularization, a sparse or low-rank metric matrix is learnt through a gradient descent procedure. The proposed algorithm is the first to combine kernel regression with sparse metric learning. KR$\_$SML is evaluated on $19$ benchmark data sets for regression. Besides, it is also applied to forecasting short-term traffic flows. For comparison purpose, three related kernel regression algorithms KR, KR\_PCA and MLKR are also employed to serve as base lines. Two widely-used criterions including the root mean square error $RMSE$ and mean absolute relative error $MARE$ are adopted to compare the performance of the four kernel regression algorithms. Experimental results of KR$\_$SML on $19$ benchmark data sets reveal competitive results. KR$\_$SML gets the best performance on almost all the data sets. Especially when a sparse metric matrix is learnt, KR$\_$SML obviously outperforms the other three kernel regression algorithms. Furthermore, experiments on real data of urban vehicular traffic flows forecasting also indicate excellent results. The promising results demonstrate that KR$\_$SML is an effective and better kernel regression algorithm. It has the capability of learning a good distance metric and simultaneously remove noise in data leading to dimensionality reduction.

KR$\_$SML targets the objective of sparse metric learning directly. It improves the performance of kernel regression by learning a sparse distance metric.
In the future, developing the potential of KR$\_$SML in other domains is our pursuit.

\section*{Acknowledgements}

This work is supported in part by the National Natural Science Foundation of China under Project 61075005, and the Fundamental
Research Funds for the Central Universities.

\end{document}